\documentclass{article}

\usepackage{microtype}
\usepackage{graphicx}
\usepackage{subfigure}
\usepackage{booktabs} 
\usepackage{amssymb}
\usepackage{pifont}
\usepackage{amsmath}
\usepackage{caption}
\usepackage[dvipsnames]{xcolor}
\newcommand{\cmark}{\ding{51}}%
\usepackage{array}
\usepackage{xcolor}
\usepackage{setspace}

\usepackage{hyperref}

\usepackage[arxiv]{mlsys2023}

\newcommand{\devashree}[1]{\textcolor{blue}{Dev: \em #1 }}

\begin{document}

\twocolumn[
\mlsystitle{PerfSAGE: Generalized Inference Performance Predictor for Arbitrary Deep Learning Models on Edge Devices}



\mlsyssetsymbol{equal}{*}

\begin{mlsysauthorlist}
\mlsysauthor{Yuji Chai}{harvard}
\mlsysauthor{Devashree Tripathy}{harvard}
\mlsysauthor{Chuteng Zhou}{arm}
\mlsysauthor{Dibakar Gope}{arm}
\mlsysauthor{Igor Fedorov}{arm}
\mlsysauthor{Ramon Matas}{arm}
\mlsysauthor{David Brooks}{harvard}
\mlsysauthor{Gu-Yeon Wei}{harvard}
\mlsysauthor{Paul Whatmough}{harvard,arm}
\end{mlsysauthorlist}

\mlsysaffiliation{harvard}{John A. Paulson School of Engineering and Applied Sciences, Harvard University, Boston, Massachusetts, United States}
\mlsysaffiliation{arm}{ARM Inc., Boston, Massachusetts, United States}

\mlsyscorrespondingauthor{Yuji Chai}{yuc927@g.harvard.edu}

\mlsyskeywords{Runtime Performance Prediction, Graph Neural Network, Deep Learning, Model Architecture}

\vskip 0.3in

\begin{abstract}
The ability to accurately predict deep neural network (DNN) inference performance metrics, such as latency, power, and memory footprint, for an arbitrary DNN on a target hardware platform is essential to the design of DNN based models. 
This ability is critical for the (manual or automatic) design, optimization, and deployment of practical DNNs for a specific hardware deployment platform.
Unfortunately, these metrics are slow to evaluate using simulators (where available) and typically
require measurement on the target hardware.

This work describes \texttt{PerfSAGE}, a novel graph neural network (GNN) that predicts inference latency, energy, and memory footprint on an arbitrary DNN TFlite graph~\cite{TFLite}.
In contrast, previously published performance predictors can only predict latency and are restricted to pre-defined construction rules or search spaces. 
This paper also describes the \texttt{EdgeDLPerf} dataset of 134,912 DNNs randomly sampled from four task search spaces and annotated with inference performance metrics from three edge hardware platforms.
Using this dataset, we train \texttt{PerfSAGE} and provide experimental results that
demonstrate state-of-the-art prediction accuracy 
with a Mean Absolute Percentage Error of $<$5\% across all targets and model search spaces. 
These results: 
(1) Outperform previous state-of-art GNN-based predictors~\cite{dudziak2020brp}, 
(2) Accurately predict performance on accelerators (a shortfall of non-GNN-based predictors \cite{zhang2021nn}),
and (3) Demonstrate predictions on arbitrary input graphs without modifications to the feature extractor.
\end{abstract}

]

\printAffiliationsAndNotice{} 



\pagenumbering{arabic}
\pagestyle{plain}
\setlength{\footskip}{40pt}

\section{Introduction}

Deep neural networks (DNNs) \cite{schmidhuber2015deep} 
are the state-of-the-art approach for
numerous tasks in the fields of computer vision, natural language processing and beyond.
Due to their subsequent popularity,
DNN inference workloads are routinely deployed on a wide range of hardware platforms, ranging from warehouse-scale computers in data centers to embedded microcontrollers in consumer edge devices.
These diverse 
hardware platforms each impose a wide variety of practical deployment constraints,
but 
tend to be 
most stringent on edge computing platforms, where inference latency, energy and memory are often severely constrained~\cite{fedorov2019sparse,banbury2021micronets}.
Therefore, application developers targeting edge platforms need to not only consider DNN accuracy, but must also meet inference latency, energy, and memory constraints.


DNN architectures are conventionally developed through trial-and-error experimentation to find a network (and training recipe) that achieves high accuracy on the target task.
This development process is further complicated when inference performance is also critical, at which point it
becomes a complex balancing act between accuracy, latency, energy consumption, and memory footprint.
These metrics conceptually present trade-offs; e.g., a larger model will typically achieve higher accuracy, but will require more memory and have longer inference latency.
However, in practice, meeting all the hardware constraints while maximizing accuracy is undeniably a very challenging hyperparameter optimization task.
To help address this challenge, 
so-called neural architecture search (NAS) techniques have been widely proposed, which employ search algorithms to automate the discovery of DNNs that meet these overlapping constraints~\cite{zoph2016neural, ProxylessNAS, liu2018darts, white2019bananas, SpArSe}.

For both manual and NAS approaches to developing DNN workloads, we require the ability 
to quickly evaluate 
inference performance metrics on the target hardware platform.
Although DNN inference runtime metrics can simply be 
measured directly on the target hardware platform, there are a number of limitations.
Firstly, this requires convenient access to the hardware, which might not be available.
Even if the hardware is available, some metrics, such as energy, are often challenging to measure.  
Secondly, 
in the Neural Architecture Search (NAS) 
setting, it is often necessary that the inference model itself be differentiable ~\cite{cai2018proxylessnas}.
NAS may also necessitate scaling to a very large throughput in performance measurement, which might require a prohibitive number of physical hardware boards.
Finally, on a typical edge SoC platform, there are usually multiple heterogeneous compute targets to consider, e.g., CPU, GPU \cite{4710975} and NPU \cite{esmaeilzadeh2012neural}.
Finding the optimal target for deployment may require benchmarking on multiple targets, which becomes slow and laborious.
A \textit{learned inference performance predictor} trained on benchmark data overcomes these limitations, providing quick and scalable performance estimation.

An ideal inference performance predictor should have the following characteristics. 
The predictor should 
generalize well to diverse input DNN graphs to deal with the inevitable evolution of network connectivity and operators.
It should predict not only latency but also 
energy and memory footprint\footnote{Memory footprint is an essential consideration for tinyML applications on embedded microcontrollers~\cite{fedorov2019sparse}}.
A single predictor model architecture should achieve high accuracy when trained on data
from a variety of hardware targets.
To date, learned DNN inference performance predictors have drawn growing research interest, with previously published 
work~\cite{zhang2021nn, wen2020neural, dudziak2020brp, kaufman2021learned, gao2021runtime}
demonstrating feasibility, but failing to meet all goals.

This paper describes \texttt{PerfSAGE}, a novel graph neural network (GNN) -based performance prediction model for DNN inference (Figure \ref{fig:overview}). 
The major advantage of a GNN feature extractor is that it can directly process arbitrary DNN graphs, generalizing well to new graph connectivity structures.
The operator-level aggregation function is trainable without requiring any hand-tuned aggregation methods used in prior work~\cite{wen2020neural, dong2018ppp},
which are time-consuming to develop and limit generalization.
The GNN model in \texttt{PerfSAGE} can even accurately model complex graph-level interactions, dictating accelerator performance.

Learned performance predictors require \textit{data} with which to train and evaluate accuracy.
We collected an open source dataset called \texttt{EdgeDLPerf}, which allows us to train and evaluate performance predictors across different DNN search spaces and hardware targets.
This is in contrast to prior works using NASBench201~\cite{dong2020bench},
which only considers convolutional neural networks (CNNs)~\cite{alexnet} for image classification.
\texttt{EdgeDLPerf} comprises models randomly sampled from four distinct DNN design spaces (Section~\ref{section:bd}), for typical edge perception tasks, with each model labeled with benchmarked inference performance metrics from three common hardware platforms.
Extending diversity beyond NASBench201, \texttt{EdgeDLPerf} models range from 7 operators, all the way to models with 1.2K operators, 
with a range of layer counts, operator types, and connectivity.
We also include recent vision transformer~\cite{xiao2021early} models.

Our experimental results (Section~\ref{section:evaluation}) demonstrate that \texttt{PerfSAGE} 
not only provides state-of-the-art accuracy
when trained \textit{individually} on each of the four tasks ($<$4.73\% MAPE), 
but it can also be trained such that a \textit{single model} achieves high accuracy on all four design spaces ($<$5.90\% MAPE).
The generalization capability 
is due to the categorical feature encoding architecture, hybrid loss, and dataset up-sampling techniques (Section~\ref{section:md}). 
In contrast, we show that prior performance predictors do not perform as well 
on our more diverse dataset.  
We further summarize the contributions of this paper below:

\begin{figure}[t]
\centering
\includegraphics[width=\linewidth]{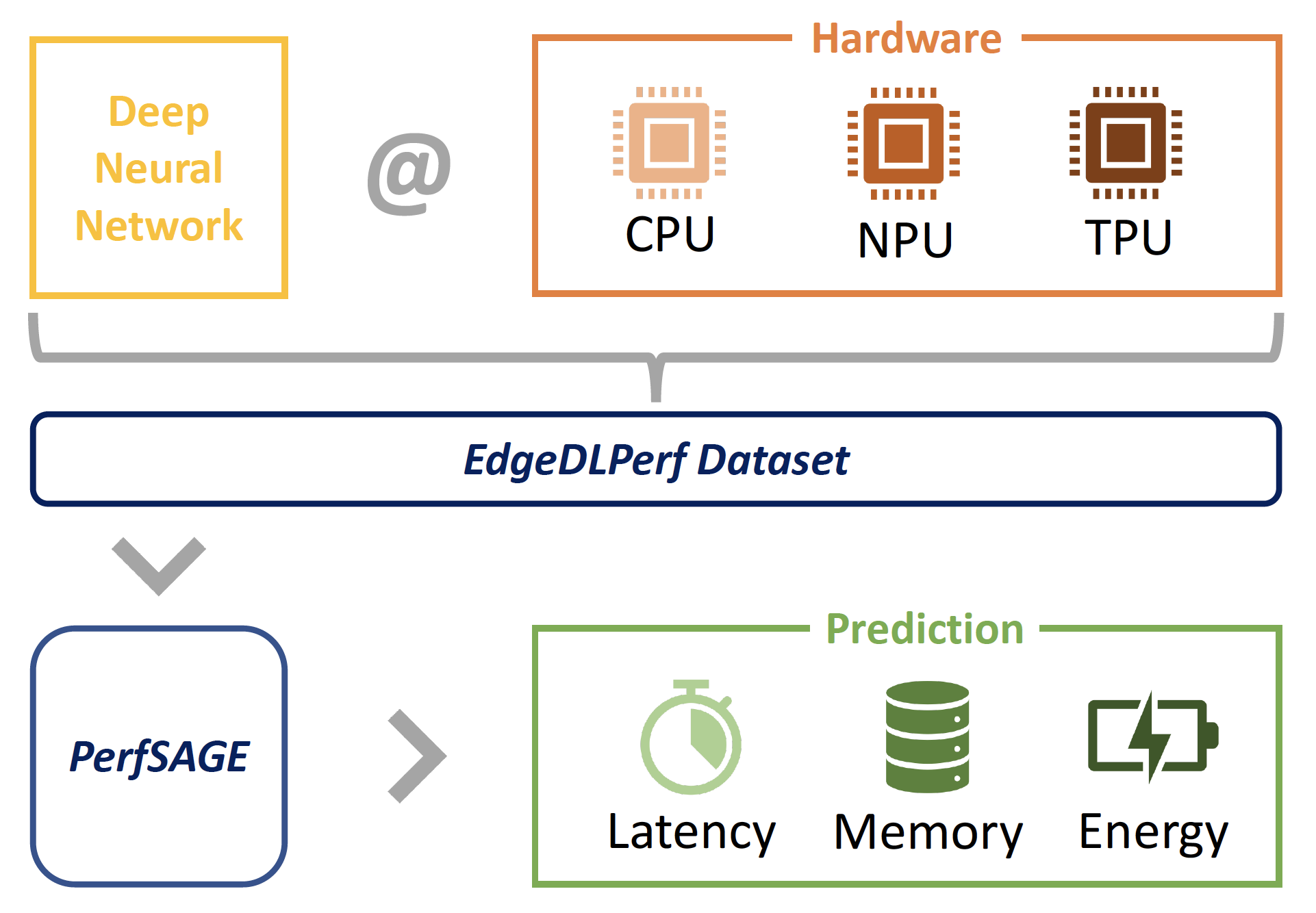}
\caption{Overview of \texttt{PerfSAGE} and \texttt{EdgeDLPerf}.}
\vspace*{-1.5em}
\label{fig:overview}
\end{figure}

\begin{enumerate}

\item \textbf{Accurate performance predictions on arbitrary DNNs without feature extractor tuning.
} 
The 
input DNN is not constrained to a specific DNN architecture construction rule (\textit{cf.}
Eagle~\cite{dudziak2020brp}), nor does it need 
tuning of the predictor model design (\textit{cf.} 
nn-Meter \cite{gao2021runtime}). 

\item \textbf{The same architecture can be trained to predict inference latency, energy, and memory with no modification. 
} 
While prior work only predicts latency, energy and memory are essential for edge device deployment.

\item \textbf{
State-of-the-art accuracy ($<$5\% MAPE)
on all prediction targets from various hardware platforms.
} 
We show that \texttt{PerfSAGE} outperforms Eagle \cite{dudziak2020brp} and nn-Meter \cite{gao2021runtime} 
on all prediction targets,
while also addressing the shortcoming of nn-Meter, which predicts poorly
on inference performance targets collected from hardware accelerators. 

\item \textbf{
An open-source dataset for evaluating DNN inference performance predictors on edge platforms.
}
\texttt{EdgeDLPerf} consists of $\sim$135K
models from four different DNN backbones, labeled with benchmark performance on three different hardware platforms.

\end{enumerate}

\begin{table*}[th!]
\centering
\small
 \captionsetup{justification=centering}
 \caption{Comparison of related works in inference performance prediction for Deep Learning models against \texttt{PerfSAGE}.}
 \label{table:taxonomy}
{\renewcommand{\arraystretch}{1.25}
 \resizebox{\textwidth}{!}{
 \begin{tabular}{|l||c|c|c|c|c|} 
 \hline
  & \textbf{LPM-TPU} & \textbf{NP-NAS} & \textbf{BRP-NAS (Eagle)}   & \textbf{nn-Meter}  & \textbf{\texttt{PerfSAGE}}\\
  & \cite{kaufman2021learned} &  \cite{wen2020neural} & \cite{dudziak2020brp}   & \cite{zhang2021nn} & \textbf{(This Work)}\\
 \hline
 \hline 
 \textbf{Hardware Targets}  & TPU &  -- & CPU, GPU, Edge-GPU  & CPU, GPU, VPU & CPU, NPU, Edge-TPU\\
 \hline 
 \textbf{Predicted Perf. Metrics} & Latency  & Accuracy & Latency, Acc-Rank & Latency &  Latency, Energy, Memory \\
 \hline
 \textbf{Dataset}  & Proprietary &  NAS-Bench & NAS-Bench  & Proprietary &  EdgeDLPerf \\
 \hline
 \textbf{DNN Architectures} & CNN, LSTM, RNN &  CNN & CNN & CNN &  CNN, Transformer \\
 \hline
 \textbf{Arbitrary DNN input w/o}  & & & & & \\
 \textbf{changing the predictor} &  \cmark &  &  & \cmark \, (\textit{Partial}) & \cmark \\
 \hline 
 \textbf{Good accuracy across} & & & & & \\
 \textbf{diverse hardware targets} & &  &  & & \cmark \\
 \hline 
 \textbf{Predict different metrics w/} & & & & & \\
 \textbf{the same predictor design} & &  & \cmark & & \cmark \\
 \hline 
 \end{tabular}
 }} \quad
\end{table*}

\section{Background and Related Work}


\subsection{Graph Neural Networks}
\label{section:gnn}

Graph neural networks (GNN) \cite{ZHOU202057} excel at tasks that require contextualized information to be extracted from a graph structure, such as is found in traffic prediction, physics simulation, and recommendation systems.
There are three kinds of prediction tasks on graphs: graph-level, node-level, and edge-level.  
A graph-level task predicts the properties of the whole graph; a node-level task predicts the properties of each node in a graph; and an edge-level task predicts the properties or presence of edges within a graph. 
GNNs solve all three levels of prediction problems.

Other DNN models such as CNNs 
and recurrent neural networks (RNNs) \cite{you2018graphrnn} can also operate on
However, they are inefficient at handling graphs 
with irregularity from unbalanced node neighbors. 
Graph Convolutional Networks (GCNs) \cite{kipf2016semi} and Graph Attention Networks (GATs) \cite{velickovic2017graph, goodfellow2014generative} are common models from the GNN family previously used for runtime-performance prediction. 
To remove the limitations of adhering to a specific graph layer, we use Spektral GNN \cite{you2020design}, which 
has 
a modularized GNN implementation and standardized evaluation.
This allows 
for convenient 
use of different GNN layers. 


DNN inference performance prediction 
is a form of 
generalized graph-level prediction.
This begins with the embedding of each node, which is then aggregated to produce a final general prediction. In a DNN model, hidden layers or operators represent nodes, whereas intermediate tensors between two layers or operators represent edges. To reduce predictor complexity, \texttt{PerfSAGE} encodes edge features as additional features for each node. As described in Section~\ref{section:md}, our edge aggregator uses the same set of parameters across a DNN model to combine neighboring edges into node embeddings. The final graph-level embedding is generated by aggregating all node embeddings. 

\subsection{Inference Performance Prediction}

DNN inference performance prediction is an active area of research. 
Researchers have recently focused on the use of learned performance predictors, which provide greater accuracy and flexibility than traditional methods.
These studies predict the runtime performance of deep learning models as well as other types of software. For example, DeepPerf \cite{ha2019deepperf} is primarily intended for analyzing the runtime performance of a C program in order to improve performance. There are prior works that are similar to this one. We will focus our discussion on previous work on runtime performance prediction for deep learning models.
Table \ref{table:taxonomy} lists prior works that are closely related.

\paragraph{DNNPerf}~\citet{gao2021runtime} propose a custom GAT based node-edge encoder predictor of GPU training time and memory consumption. It supports diverse frameworks and device options. Its predictor is trained using a proprietary dataset that includes LeNet \cite{lecun1998gradient}, ResNet-V1 \cite{he2016deep}, Inception-V3 \cite{szegedy2016rethinking}, Vanilla RNN \cite{rumelhart1986learning}, LSTM \cite{graves2012long}) and tested on AlexNet \cite{krizhevsky2012imagenet}, VGG \cite{simonyan2014very}, OverFeat \cite{sermanet2013overfeat}, ResNet-V2 \cite{he2016deep}, and GRU \cite{cho2014learning}. The DNNPerf model focuses primarily on training on cloud-based GPUs. Our focus is on predicting inference performance on diverse edge devices. DNNPerf has been excluded from Table \ref{table:taxonomy} due to space limitations. 

\paragraph{LPM-TPU}~\citet{kaufman2021learned} describe a learned performance model for XLA tensor programs on TPU \cite{XLA}. The computational graph of the XLA program is decomposed into smaller sub-graphs, and separate instances of the model are trained for distinct optimization tasks such as tile size selection and operator fusion. They generalize well for programs that have some similarity to the training set. However, their model requires a compiled XLA format model as input, which is not easily accessible to users of various hardware platforms. In contrast, \texttt{PerfSAGE} directly takes a saved model as input and shows consistent performance across multiple platforms.

\paragraph{NP-NAS}~\citet{wen2020neural} describe an approach to find a good network candidate for prediction accuracy using the top-K strategy. Their proposed neural predictor generalizes well to models from the NASBench-101 \cite{ying2019bench}. However, their predictor still relies on a hand-tuned encoding method for a well-defined design space. 

\paragraph{BRP-NAS (Eagle)}\citet{dudziak2020brp} present a neural architecture search method to find the most performant model under a latency constraint. 
By exploiting a GCN-based predictor, they shrink the search space and accelerate the search process. 
The Eagle predictor used in BRP-NAS \cite{dudziak2020brp} is designed to only support models constructed from NASBench201 architectures. 
It exploits the NASBench201 construction rule to reduce the complexity of input model architectures.
If the construction rule of the model architecture is previously unknown, the accuracy suffers significantly, as shown in Table \ref{table:ete}. 
\texttt{PerfSAGE}, on the other hand, performs well despite the fact that it does not take the construction rule as input.

\paragraph{nn-Meter}~\citet{zhang2021nn} and \textbf{NetAdapt}~\citet{NetAdapt} both predict the total latency of a DNN on an edge device, by making estimating latency for each layer 
and subsequently adding all the layer latencies together.
\mbox{NetAdapt} only takes models with well-defined architecture rules. While nn-Meter claims to support arbitrary CNN models, their fusion rule still poses some limitations on input model architecture, based on our testing results. 
Furthermore, their methods are not easily portable to different hardware platforms. 
The predictor performance of nn-Meter, for example, drops significantly for data collected from a deep learning accelerator.  
This shortcoming has been overcome by \texttt{PerfSAGE}, as discussed in Section \ref{section:s&g}.



\section{EdgeDLPerf Dataset}
\label{section:bd}


To capture the diverse landscape of modern deep learning models, we collected a benchmark dataset for edge deployments (\texttt{EdgeDLPerf}).
It includes a collection of deep learning models with their benchmarked runtime performance metrics on various edge hardware platforms. 
These models are diverse in two major ways.
Firstly, \texttt{EdgeDLPerf} contains models across many different applications. 
Section~\ref{model_gen} describes how \texttt{EdgeDLPerf} is generated through a sampling process, and then we provide details on how the design space is constructed for different applications (Section~\ref{section:ddlm}).
Secondly, \texttt{EdgeDLPerf} contains diverse prediction labels from benchmarking runs on various hardware platforms. 
The diversity of hardware platforms and runtime performance metrics, further discussed in Section \ref{section:dpt}, showcases  \texttt{PerfSAGE}'s generalization capability. 
The benchmarking strategy is discussed in Section \ref{section:bm}. 
Therefore, compared to datasets derived from NASBench-101~\cite{ying2019bench} and NASBench-201~\cite{dong2020bench}, \texttt{EdgeDLPerf} covers more application domains (e.g., keyword spotting) and novel model architectures (e.g., vision transformers). It also contains multi-target hardware performance metrics as labels and is in the inference-ready TFLite model format. 

\subsection{Sampling Random Models from a Design Space}
\label{model_gen}
We leverage the search space construction  and sampling mechanism in differentiable neural architecture search (NAS)~\cite{fedorov2022udc} to sample the models forming \texttt{EdgeDLPerf}. 
During NAS, a supernet model is defined with many searchable options for the network architecture. 
Each option is associated with a categorical random variable defined by its underlying probabilities. 
Sampling from these option random variables gives a sub-graph in the supernet, which is one sampled model from the search space. 
In NAS, the sampling is usually performed with a differentiable Gumbel-softmax type differentiable sampler~\cite{jang2016categorical}. 
In this work, our goal is to generate a large collection of different models from the search space. Therefore, we use a uniform sampling strategy which directly samples from uniform categorical random variables. 
We further make the assumption that all the options are independent from each other. 
A sampled model graph is converted into TFLite model format with the TFLite converter. 
All the models are quantized to INT8 for both weights and activations with the TFLite post-training quantization tool since it is the most relevant format for edge deployment. 
The model weights are not trained and are from random initializations. 
The sampled TFLite models are split into $80\%$ training data and $20\%$ test data.

\begin{table}[th!]
\centering
\small
 \caption{
 Summary of EdgeDLPerf dataset characteristics and types of operators used in each design space.}
 \label{table:operand}
{\renewcommand{\arraystretch}{1.2}
\resizebox{\columnwidth}{!}{
 \begin{tabular}{|>{\bfseries}l||c|c|c|c|} 
 \hline
                        & \textbf{\emph{CNN-}} & \textbf{\emph{CNN-}} & \textbf{\emph{CNN-}}  & \textbf{\emph{ViT-}} \\
        & \textbf{\emph{CIFAR10}} & \textbf{\emph{ImageNet}} & \textbf{\emph{KWS}}  & \textbf{\emph{CIFAR100}} \\
 \hline 
 \hline 
 \multicolumn{5}{|c|}{\bf{Characteristics}} \\
 \hline 
 \#Samples              & 39,997   & 79,998 & 4,917 & 10,000 \\
 \hline
 \#Operators            & 13-70   & 16-77 & 7-21 & 59-994 \\
 \hline
 \#Tensors              & 28-145   & 39-182 & 16-51 & 97-1264 \\
 \hline
 Size (MB)              & 0.09-18.25   & 1.65-20.54 & 0.04-5.05 & 1.86-53.98 \\
 \hline 
 \hline 
 \multicolumn{5}{|c|}{\bf{Operators}} \\
 \hline
 DENSE        & & & & \cmark    \\
 \hline
 CONV 2D                & \cmark    & \cmark    & \cmark    & \\
 \hline 
 DEPTHWISE      & \cmark    & \cmark    & \cmark    & \\
 \hline 
 AV POOL 2D        & \cmark    & \cmark    & \cmark    & \\
 \hline 
 PAD                    & \cmark    & \cmark    & & \\
 \hline 
 ADD                    & \cmark    & \cmark    & & \cmark    \\
 \hline 
 MUL                    & & & & \cmark    \\
 \hline 
 POW                    & & & & \cmark    \\
 \hline 
 TANH                   & & & & \cmark    \\
 \hline 
 SOFTMAX                & & & & \cmark    \\
 \hline 
 RESHAPE                & \cmark    & \cmark    & \cmark    & \cmark    \\
 \hline 
 TRANSPOSE              & & & & \cmark    \\
 \hline 
 STR SLICE          & & & & \cmark    \\
 \hline 
 CONCAT          & & & & \cmark    \\
 \hline 
 \end{tabular}
 }} \quad
\end{table}

\subsection{Design Spaces for Edge Deep Learning}
\label{section:ddlm}
Recent progress in deep learning has seen the appearance of many different model architectures. 
Architecture differences may come from different domains of applications, accuracy requirements, latency requirements, and so on. 
In order to target a wide range of applications and the associated models, we picked four different model design spaces, corresponding to four different tasks, and sampled a collection of models from each of them using the sampling method described in Section \ref{model_gen}. 
Together, they form the \texttt{EdgeDLPerf} dataset.

\begin{itemize}
  \item \textbf{CNN-CIFAR10} Convolutional Neural Networks designed for image classification task targeting the CIFAR10 dataset. 
The design space for CNN-CIFAR10 is built with the same building blocks as CNN-ImageNet, using $3$ blocks instead of $6$. 
The searchable channel numbers are also uniformly reduced by a factor of $3$ to reflect that CIFAR10 is a smaller dataset. As a result of this, and taking into account that the CIFAR10 dataset has a lower resolution, the models from this design space are smaller and run faster.

  \item \textbf{CNN-ImageNet} Convolutional Neural Networks designed for image classification task  targeting the ImageNet dataset. 
 For this application, we have used a search space inspired by EfficientNet-EdgeTPU~\cite{gupta2020accelerator}. 
  The design space is composed of 6 independent blocks, each of them consisting of $5$ nodes choosing over inverted bottleneck, fused MBConv, and an identity function. The design space is also searching over channel numbers ($6$ different) and kernel sizes ($3$$\times$$3$ and $5$$\times$$5$) in each node, as well as optional residual connections at the block level.
  
  \item \textbf{CNN-KWS} Convolutional Neural Networks designed for audio keyword spotting targeting the Google Speech Commands datasets.
The design space for these models is derived from the one used in \citet{banbury2021micronets} by adding kernel size options, which is a design space optimized for the edge deployment scenario on commodity microcontrollers. 
  As a result, they show the smallest numerical values for the inference performance metrics and are more prone to prediction errors. 
  This design space also contains models that have non-square shaped convolutional filters.
  \begin{table}[th!]
\centering
\small
 \caption{Hardware platforms and prediction targets.}
 \label{table:hardware}
{\renewcommand{\arraystretch}{1.15}
 \begin{tabular}{|>{\bfseries}l|c|c|} 
 \hline
 & \textbf{HW Platform} & \textbf{Prediction Targets} \\
 \hline
 \hline
 CPU & Quad-Core Cortex-A57  & Latency, Energy \\
 \hline 
 NPU & Arm Ethos-U NPU            & Latency, SRAM Usage\\
 \hline 
 TPU & Google EdgeTPU     & Latency$^\dagger$ \\
 \hline 
 \end{tabular}} \quad
 \\
 \footnotesize
 $^\dagger$ EdgeTPU does not allow energy or memory measurements.
\end{table} 
  \item \textbf{ViT-CIFAR100} Vision Transformer-based models designed for image classification 
  on the CIFAR100 dataset. Since all models in this design space use transformer-based model architecture, their architectures differ significantly from other design spaces.
  The large search space for these models includes the vision transformer's main changeable dimensions, such as embedding dimension, number of heads, query/key/value dimension, MLP ratio, and network depth. This space contains a plethora of transformers with diverse structures and model complexities.
  As a result, \emph{ViT-CIFAR100} is usually the hardest search space.
\end{itemize}
To provide a better representation of models from different design spaces, we represent each model with a graph. 
Every layer or operand is considered a graph node, while every tensor between operands is 
an edge between two nodes. 
We can represent every deep learning model with its graph-level topology and characteristics. 
Table \ref{table:operand} shows the number of models sampled, the number of operands range, the number of tensors range, and the parameter count range for each design space. 
It also shows different types of operands present in each design space. 
Though the three CNN design spaces (\emph{CNN-CIFAR10}, \emph{CNN-ImageNet}, and \emph{CNN-KWS}) share similar operand types, they differ widely in their model sizes.
\emph{ViT-CIFAR100} shows a very different collection of operands. 
It also has a significantly higher number of operands with its transformer-based structures. 
As shown in Table \ref{table:operand}, \texttt{EdgeDLPerf} is composed of a wide range of deep learning models built for different applications, with extensive diversity in model graph characteristics and operand types.

\subsection{Edge Hardware Performance Labels}
\label{section:dpt}

\begin{figure*}[t!]
\centerline{\includegraphics[width=0.95\textwidth]{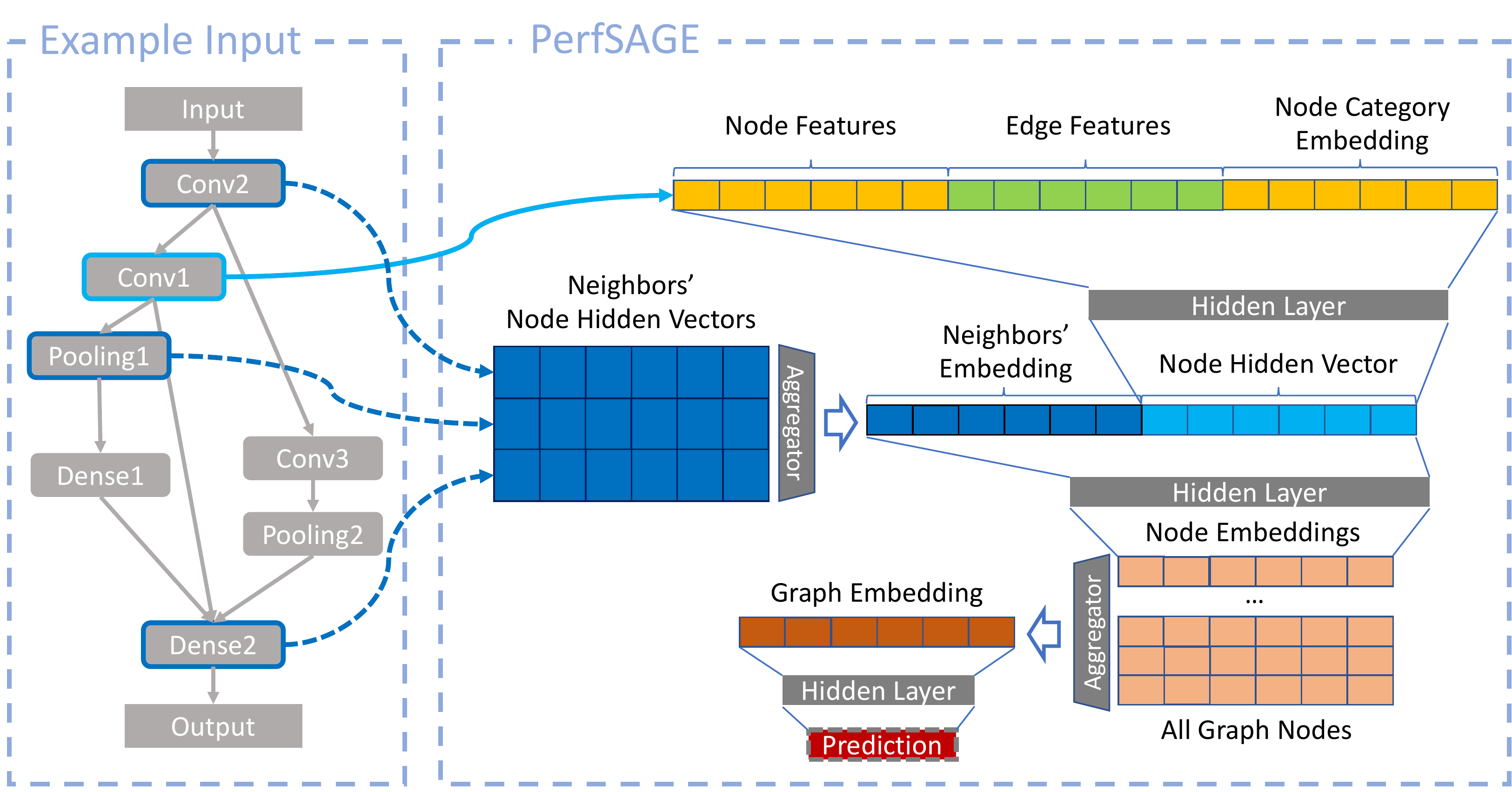}}
\vspace{-1em}
\caption{Model Architecture of \texttt{PerfSAGE}.}
\label{figure:architecture}
\end{figure*}

Different hardware targets can show wildly different performance responses to variations in model connectivity and operators.
For example, the latency and energy of models running on a CPU often show a more direct correlation with the parameter count and number of MACs. 
This correlation comes from the somewhat limited parallelism on a CPU.
On the contrary, accelerators such as the NPU~\cite{esmaeilzadeh2012neural} and TPU~\cite{jouppi2017datacenter} 
exploit massive parallelism, especially for operands such as convolution. 
Thus, results from NPU and TPU usually may not directly scale with the first-order model dimensions.
To capture this diversity, we used three different hardware targets to add labels to each model in the dataset, as shown in Table~\ref{table:hardware}.
For CPU and TPU benchmarks, we are able to directly deploy each model onto the hardware to measure performance data.
For the NPU, we collected performance from the open source simulator Vela\footnote{Vela is available at: \url{https://git.mlplatform.org/ml/ethos-u/ethos-u-vela.git/about/}}, supporting the Arm \mbox{Ethos-N} NPU.
Using Vela, 
we could also measure the memory (SRAM) usage during inference, which is not always easy to obtain by direct deployment. 
However, 
Vela does not currently 
support latency estimation for \emph{ViT-CIFAR100} due to limited operator support.

\subsection{Hardware Benchmarking Methodology}
\label{section:bm}

After generating the sampled models, we next add the measured performance labels for each model, which are the prediction targets.  
The labels are generated by measuring each model on the hardware platforms (CPU on NVIDIA TX2 Board\cite{nvidia_developer_2022}, Arm Ethos-U NPU\cite{Ethos-NPU} and Edge TPU on Coral Dev Board\cite{coral}), shown in Table~\ref{table:hardware}.
Each benchmarking run consists of two parts:
warm-up, where the inference is initially looped to warm up caches, and measurement, where the performance metrics are captured.
Random data is used for both the DNN input (batch size one) and the weight tensors.
All the measurements that we have taken are average results from several inference runs to reduce random noise during measurement. 
The power measurement is based on the average of several power readings that we took while the inference was looping.
All the hardware platforms were used in 
maximum performance mode,
providing extra cooling 
if thermal throttling was observed. 
\section{PerfSAGE Design}
\label{section:md}


\texttt{PerfSAGE} 
(Figure~\ref{figure:architecture}), aims to capture features from a diverse range of DNNs.  
To achieve this, it uses a GNN to encode arbitrarily defined input model graphs.
As 
discussed in Section~\ref{section:gnn}, GNNs are capable of inductive inference on unseen graph structures.


\subsection{Model Architecture}

Figure \ref{figure:architecture} (left) shows a hypothetical CNN input graph. A node, $i$, in the input graph represents an operator (gray boxes, e.g., \textit{Conv1}, \textit{Pooling1}, etc.), and a directed edge connecting nodes represents a tensor between the operators. 
If we consider operator \textit{Conv1} (grey node in the input graph), as an example for node $i$, the corresponding node type $c_i$ would be \textit{CONV 2D} (node type not shown in the figure).
The feature input for a node consists of three parts: node features (yellow boxes), edge features (green boxes), and node category embedding (yellow boxes). 
Node features $F_{node}(i)$ are the corresponding layer's hyperparameters, such as padding size or convolution kernel dimensions. 
The output dimensions of that node are considered as the edge features $F_{edge}(i)$. 
The node category embedding $E_{cat}(c_i)$ comes from a trainable embedding table, which stores a collection of parameter vectors for different types of node.
Only the parameter vector corresponding to the current node type will be fed into the node category embedding vector. 
Then, we concatenate the three feature vector parts and use a hidden layer to generate the \textit{node hidden vector} (light blue box) $h_{i}^{0}$, described by Equation $(1)$ below 
\begin{equation}
\resizebox{.9\hsize}{!}
{$h_{i}^{0} = \mathrm{relu}(w_{c_i}^{0} * (F_{node}(i) || F_{edge}(i) || E_{cat}(c_i)) + b_{c_i}^{0}).$}
\end{equation}
Considering only the feature inputs from node $i$, i.e., the \textit{Conv1} node, is not sufficient, as the topological structure of the entire graph is also essential to predict performance on an input DNN.
Thus, 
our approach is to incorporate the directly connected node neighbors (\textit{Conv2}, \textit{Pooling1}, and \textit{Dense2}) feature information as well. 
 
To achieve this in the example of Figure \ref{figure:architecture}, we collect neighbor node information by aggregating the \textit{node hidden vectors} (dark blue boxes) through a sum kernel to generate the \textit{neighbor embeddings}. 
The node hidden vector concatenated with neighbors' embedding is fed through eight GNN layers, before being transformed into the final Node Embedding for node $i$.
The $k^{th}$ layer of graph neural network, shown in Equation $(2)$ below, takes the previous layer's hidden vector $h_{i}^{k-1}$ of node $i$ and hidden vectors from its neighbors $\{h_{j}^{k-1}, j\in N(i)\}$ to produce hidden vector $h_{i}^{k}$, thus,
\begin{equation}
\resizebox{.9\hsize}{!}
{$h_{i}^{k} = \mathrm{relu}((h_{i}^{k-1}||\mathrm{SUM}(\{h_{j}^{k-1}, j\in N(i)\})) * w^{k} + b^{k}).$}
\end{equation}
To predict properties of the entire graph, not just individual nodes, we aggregate information from all nodes with a maximum aggregation kernel to obtain the final graph embedding vector.
This vector includes information about each node and its neighbors. From this graph embedding, the final prediction is generated through 4 fully connected layers, with a hidden dimension of 1024.

\subsection{Categorical Feature Encoding}
One of the most important features for each operator node is its category or type (e.g., \textit{CONV 2D}, \textit{FULLY CONNECTED}, \textit{ADD}, etc.). 
Not only may two nodes with different categories have completely different kinds of features, they may also have completely different meanings for the same feature. 
For example, consider the output dimension of a node.
A \textit{CONV 2D} layer would have a vastly different layer property than an \textit{ADD} operator with the same output dimension. Their computational operational count or usage of hardware resources would be completely different. 
If they nevertheless share the same set of parameters, the training process with different node types would be quite counterproductive for the node output dimension parameter.
As the type of node changes, their training gradient will point in a completely different direction.
As a result, we propose categorical feature encoding (CFE), which includes a lookup table of parameters based on the category of each node. 
As shown in Equation (1), the weight $w_{c_i}^{0}$ and bias $b_{c_i}^{0}$ of node $i$ are dependent on its type $c_i$.
Each node feature consists of the operator's hyperparameter and its operator type. 
With CFE, the node output dimensions of \textit{CONV 2D} and \textit{ADD} would be considered two different types of parameters rather than the same type.
This look-up table based method ensures that different types of node can have completely different sets of weights for their input features, even if they share some features. 
To conclude, CFE grants \texttt{PerfSAGE} the capability to adopt a different number of features as well as the same feature with a different effect on the prediction target.

\subsection{Hybrid Loss Function}
As shown in Table \ref{table:operand}, models from \texttt{EdgeDLPerf} are quite diverse (very small to very large), which results in a wide numerical range for the runtime performance metric regression problem.
The numerical value of the smallest prediction target is an order of magnitude higher than that of the largest prediction target, e.g.
latency ranges from 2ms to 34ms (Figure \ref{figure:loss-bin-imagenet}). 
Due to the wide numerical range for most metrics, percentage error is the primary concern for most potential users. 
If we always use an absolute error as the optimization target, a 1ms absolute error means an insignificant 3\% error rate for a 30ms target, but would imply a 33\% error rate on a 3ms target. 
As a result, using Mean Absolute Percentage Error (MAPE) is a more reasonable general optimization and comparison target 
and provides a relatively low error rate across the prediction target range. 
However, when the prediction target magnitude is high, MAPE alone is not enough to restrict absolute error.
So, 
we also use Mean Square Error (MSE) as a supplemental loss function to restrict the error if the numerical value for prediction is high. 
The hybrid loss function is shown in Equation~(4).
It uses an additional hyperparameter $\alpha_{MSE}$, tuned to ensure equal loss contributions from the MAPE and MSE sides, resulting in optimal results. 
The hybrid loss is defined as
\begin{equation}
\resizebox{.9\hsize}{!}{$
\mathrm{MSE}(X, Y) = \sqrt[]{(X-Y)^2}, \mathrm{MAPE}(X, Y) = \frac{|X-Y|}{Y},$}
\end{equation}
\begin{equation}
\resizebox{.9\hsize}{!}{$
\mathrm{Loss}(X, Y) = \mathrm{MAPE}(X, Y) + \alpha_{MSE} * \mathrm{MSE}(X, Y).$}
\end{equation}

\subsection{Dataset Upsampling}
It is common for  different design spaces to have non-uniformly sampled data counts. 
A design space with many data points may have a large impact on the generalized model and other design spaces, while smaller design spaces suffer from insufficient data points, as with \emph{CNN-KWS}. 
To resolve this biased distribution, we introduce dataset-based upsampling.
Depending on the size of the dataset, we ensure that models from smaller datasets appear more frequently during the training process, while those from larger datasets appear at least once. 
The upsampling ratio is determined by up-sampling the smaller dataset to have at least 10\% of all data points in the training set. 
This ensures that models from a smaller design space have a larger impact 
during the training process.
Hence, models from larger design spaces will not be ignored due to their dominance.

\section{Evaluation Results}
\label{section:evaluation}
This section presents experiments evaluating
\texttt{PerfSAGE}.
Multiple models are trained, with ablations to evaluate the effectiveness of different training methods (Section
\ref{section:as}).
We also quantitatively compare 
with 
Eagle~\cite{dudziak2020brp} and 
nn-Meter~\cite{zhang2021nn}. 
All comparison experiments use the \texttt{EdgeDLPerf} dataset (Section~\ref{section:ddlm}).

\subsection{GNN Training}
For training purposes, we randomly select $80\%$ of all data points, while the other $20\%$ are used exclusively for testing. 
Every experiment uses the same training and testing split.
All 
results are compared based on mean absolute percentage error (MAPE), which provides a general comparison metric for all numerical ranges. 
In our comparisons, we train models using two different approaches: \textit{specialized} and \textit{generalized}.
In specialized mode, the predictor model is trained for a single design space at a time, yielding four unique  models. 
In generalized mode, a single model is trained to predict on any of the design spaces at once.
The generalized error rate is generated from testing results for each design space using the same generalized model. Every model is trained on an NVIDIA A100 with a time budget of 8 hours. 
All of our models achieve convergence within that budget. 

\subsection{Ablation Study}
\label{section:as}

\begin{table}[t]
\centering
\small
 \caption{Impact of Categorical Feature Encoding (CFE) on prediction MAPE for Latency (CPU) of different design spaces.}
 \label{table:cfe}
{\renewcommand{\arraystretch}{1.15}
 \begin{tabular}{|>{\bfseries}l||c|c|c|c|} 
 \hline
  & \textbf{\emph{CNN-}}     & \textbf{\emph{CNN-}}     & \textbf{\emph{CNN-}}  & \textbf{\emph{ViT-}} \\
  & \textbf{\emph{CIFAR10}}  & \textbf{\emph{ImageNet}} & \textbf{\emph{KWS}}   & \textbf{\emph{CIFAR100}} \\
 \hline
 \hline
  & \multicolumn{4}{c|}{\textbf{\texttt{PerfSAGE} (Specialized)}} \\
 \hline
 w/o CFE & 4.800\% & 1.439\% & 7.043\% & 11.471\% \\
 \hline 
 with CFE & \textbf{2.314\%} & \textbf{1.098\%} & \textbf{3.722\%} & \textbf{4.451\%} \\
 \hline 
 \hline
  & \multicolumn{4}{c|}{\textbf{\texttt{PerfSAGE} (Generalized)}} \\
 \cline{2-5}
 \hline
 w/o CFE & 4.136\% & 4.342\% & 5.969\% & 6.505\% \\
 \hline 
 with CFE & \textbf{2.617\%} & \textbf{1.263\%} & \textbf{5.847\%} & \textbf{4.798\%} \\
 \hline 
 \end{tabular} } \quad
\end{table}

In this section, we use an ablation to show the effectiveness of our proposed unique techniques,
including categorical feature encoding,  hybrid loss function, and dataset up-sampling. 

\subsubsection{Categorical Feature Encoding (CFE)}
CFE helps 
provide different parameters for different types of node with similar node features. 
Different parameters allow 
to differentiate between different nodes and provide different prediction results, even if their node features are similar. 
CFE is tested with specialized models and the generalized model with latency (CPU) as the prediction target. 
Table \ref{table:cfe} shows the results for both scenarios, where we see
CFE is crucial for both specialized and generalized models.
For specialized models, \emph{ViT-CIFAR100} shows the most significant improvements due to CFE. 
CFE is essential for understanding how different operands can have completely different properties while having similar hyperparameters.
For the other three CNN design spaces, latency is usually dominated by their convolution layers.
As a result, 
performance does not completely deteriorate, even if its parameters do not differentiate between different types of operators. 
The models in \emph{ViT-CIFAR100} have no dominant operator type; hence, the larger improvement. 
In the generalized model comparison, tests without CFE show that it is hard to understand differences between different layer types.
This even further reduces performance on \emph{CNN-ImageNet}, which performs well in specialized tests.
This comparison of specialised models and generalised models demonstrates the critical importance of CFE, especially as the model design space becomes increasingly diverse.

\subsubsection{Hybrid Loss Function}

\begin{figure}[t!]
\centerline{\includegraphics[width=0.5\textwidth]{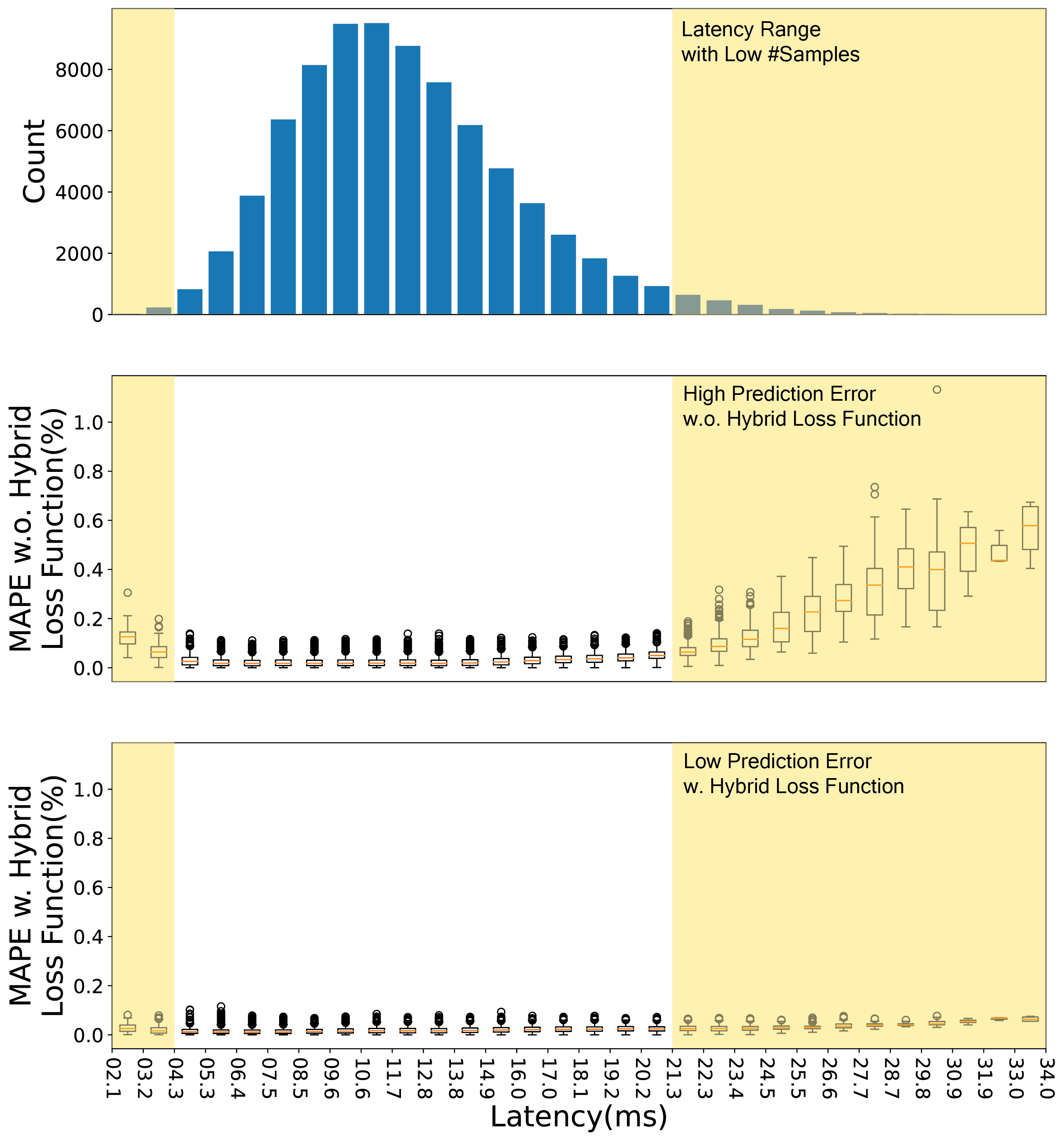}}
 \vspace{-3.5mm}
\caption{Detailed box plot of Mean Absolute Percentage Error (MAPE) in different bins, sorted by numerical value of prediction targets. The reported results come from predictions for CPU inference latency on \emph{CNN-ImageNet}.
}
\vspace{-3.5mm}
\label{figure:loss-bin-imagenet}
\end{figure}

\begin{figure}[t!]
\centerline{\includegraphics[width=0.5\textwidth]{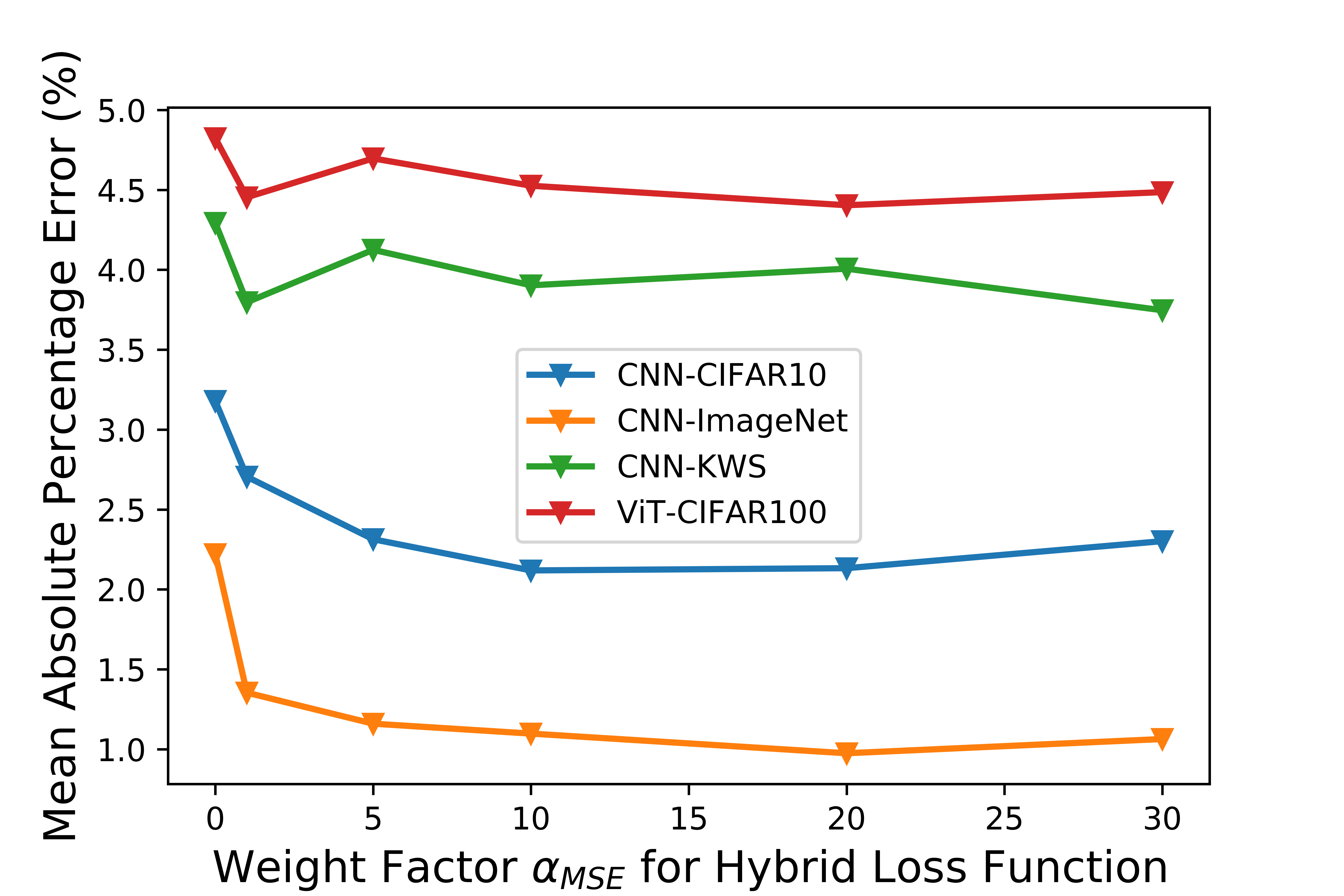}}
\vspace{-1.55 mm}
\caption{Impact of weight factor for Hybrid Loss Function on prediction MAPE for latency (CPU) on \texttt{EdgeDLPerf} dataset.}
\vspace{-6.4 mm}
\label{figure:loss-compare}
\end{figure}

To understand how the introduction of the hybrid loss function helps with the final test error, we compared test MAPE for latency prediction on CPU using the \emph{CNN-ImageNet} design space. 
In Figure \ref{figure:loss-bin-imagenet}, we generate the prediction error rate for different latency ranges across the entire dataset.
The plot is separated into two parts. 
The left plot shows the test MAPE 
with just the simple MAPE 
loss function, while the right plot shows the results for using a hybrid loss function. 
The first row of the plot shows the histogram of all model latencies, and the following two rows show the MAPE 
and the 
MAPE across the entire latency range. 
Without the hybrid loss function, there is a clear rise in error for larger latency ranges, and the MAPE also struggles in latency bins where there are fewer data points.
This kind of error in the higher latency range results from MAPE and cannot easily limit MAPE when the magnitude of latency is higher. 
Hybrid loss can easily limit this by imposing extra limits on absolute errors instead of percentage errors. 
As shown by the right plot, the higher error rate in the larger latency range is successfully limited by the MSE loss. 
It also helps to reduce the loss for latency ranges where the model density is much lower.
In Figure \ref{figure:loss-compare}, we also investigate the effect of the weight factor, $\alpha_{MSE}$, on MSE loss. 
As the weight increases, the MAPE loss decreases and then plateaus.
If the $\alpha_{MSE}$ is too high, it will cause some unnecessary training noise. 
We empirically chose the $\alpha_{MSE}$ to be 10 for 
our optimal training configurations. 

\subsubsection{Dataset UpSampling}

Dataset upsampling is critical
to the generalization of \texttt{PerfSAGE} on
datasets with biased model density.
We compared 
model performance with and without upsampling, 
with all other hyperparameters and training methods the same.
We upsample the \emph{CNN-KWS} by 3$\times$ and the \emph{ViT-CIFAR10} by 2$\times$ to reach 10\% of all training examples. 
From Table \ref{table:up-sample}, it is evident that up-sampling could greatly improve 
performance with a biased dataset distribution. 
The improvement is the most significant on datasets with a smaller number of data points. 
At the same time, upsampling does not hurt 
performance on datasets with large model count.

\begin{table}[t]
\centering
\small
 \caption{Impact of Dataset Upsampling (DUS) on prediction MAPE for latency (CPU) of different design spaces.}
 \label{table:up-sample}
    \renewcommand{\arraystretch}{1.15}\resizebox{\columnwidth}{!}{
     \begin{tabular}{|!{\bfseries}c||c|c|c|c|} 
     \hline
      & \multicolumn{4}{c|}{\textbf{\texttt{PerfSAGE} (Generalized)}} \\
      \cline{2-5}
      & \textbf{\emph{CNN-}}    & \textbf{\emph{CNN-}}      & \textbf{\emph{CNN-}}  & \textbf{\emph{ViT-}} \\
      & \textbf{\emph{CIFAR10}} & \textbf{\emph{ImageNet}}  & \textbf{\emph{KWS}}   & \textbf{\emph{CIFAR100}} \\
     \hline
     \textbf{w/o DUS} & 3.080\% & 1.592\% & 8.475\% & 6.710\% \\
     \hline 
     \textbf{w/ DUS} & \textbf{2.617\%} & \textbf{1.263\%} & \textbf{5.847\%} & \textbf{4.798\%} \\
     \hline 
     \end{tabular}
    } \quad
  \vspace{-3mm}
\end{table}

\subsection{Overall Results \& Comparison with Related Work}

\begin{table*}[t]
\centering
\small
 \caption{\texttt{PerfSAGE} Mean Square Percentage Error (MAPE) when predicting various targets from the \texttt{EdgeDLPerf} dataset. Results from both specialized and generalized models are shown. For comparison, we also show MAPE results for  Eagle~\cite{dudziak2020brp} when used to predict various targets from our \texttt{EdgeDLPerf} dataset. Eagle is primarily designed for NASBench201, a well defined design space with detailed architecture rules. Thus, it struggles with a more diverse collection of models. Due to its limited generalization capability, only specialized results are shown. The latency results for \emph{ViT-CIFAR100} on NPU are not included due to lack of transformers support on the ARM Ethos-U NPU.
}
 \label{table:ete}
{\renewcommand{\arraystretch}{1.15}
 \begin{tabular}{|>{\bfseries}c||c|c|c|c|c|} 
 \hline
 & \textbf{Latency (CPU)} & \textbf{Energy (CPU)} & \textbf{Latency (NPU)} & \textbf{Memory (NPU)} & \textbf{Latency (TPU)} \\ 
 \hline
 \hline
 \multicolumn{6}{|c|}{\textbf{\texttt{PerfSAGE} (Specialized)}} \\
 \hline
 \emph{CNN-CIFAR10}    & \textbf{2.314\%} & \textbf{3.164\%} & \textbf{2.103\%} & \textbf{2.758\%} & \textbf{2.757\%} \\ 
 \hline 
 \emph{CNN-ImageNet}   & \textbf{1.098\%} & \textbf{1.131\%} & \textbf{1.352\%} & 3.599\% & 2.360\% \\ 
 \hline 
 \emph{CNN-KWS}        & \textbf{3.722\%} & 4.455\% & 3.574\% & \textbf{3.248\%} & \textbf{3.055\%} \\ 
 \hline 
 \emph{ViT-CIFAR100}   & \textbf{4.451\%} & \textbf{4.633\%}    & --$^\dagger$    & \textbf{0.701\%} & \textbf{4.730\%} \\ 
 \hline 
 \hline 
 \multicolumn{6}{|c|}{\textbf{\texttt{PerfSAGE} (Generalized)}} \\
 \hline
 \emph{CNN-CIFAR10} & 2.617\% & 3.694\% & 3.157\% & 3.625\% & 2.816\% \\ 
 \hline  
 \emph{CNN-ImageNet}& 1.263\% & 1.474\% & 2.115\% & \textbf{3.423\%} & \textbf{1.345\%} \\ 
 \hline  
 \emph{CNN-KWS}     & 5.847\% & \textbf{4.259\%} & 6.390\% & 4.952\% & 5.115\% \\ 
 \hline  
 \emph{ViT-CIFAR100}   & 4.798\% & 5.901\%  & --$^\dagger$    & 1.580\% & 4.830\% \\ 
 \hline 
 \hline
 \multicolumn{6}{|c|}{\textbf{Eagle (Specialized)}} \\
 \hline
 \emph{CNN-CIFAR10}     & 41.138\% & 38.482\% & 36.388\% & 39.625\% & 85.079\% \\  
 \hline 
 \emph{CNN-ImageNet}    & 18.080\% & 16.745\% & 15.412\% & 21.908\% & 96.921\% \\  
 \hline 
 \emph{CNN-KWS}         & 36.142\% & 39.464\% & 35.131\% & 24.575\% & 57.823\% \\  
 \hline 
 \emph{ViT-CIFAR100}    & 46.203\% & 52.256\%  & --$^\dagger$    & 50.503\% & 91.550\% \\ 
 \hline 

 \end{tabular}} \quad \\
 \tiny
 $^\dagger$ Missing operator support for ViT models in Vela, which prevents characterization.
\end{table*}

\begin{table}[t]
\centering
\small
 \caption{Mean Square Percentage Error (MAPE) on predicting various targets from EdgeDLPerf dataset using nn-Meter~\cite{zhang2021nn}. \texttt{PerfSAGE} results are also included for comparison. The nn-Meter design is primarily used for latency prediction and is not suitable for other inference performance metrics. Moreover, it is not flexible enough to support all models in our dataset. }
 \label{table:nnm_comparison}
{\renewcommand{\arraystretch}{1.15}
\resizebox{\columnwidth}{!}{
 \begin{tabular}{|>{\bfseries}c||c|c|c|} 
 \hline
 & \multicolumn{3}{c|}{\textbf{Latency (CPU)}}\\ 
 \hline
 \hline
              &          & \textbf{PerfSAGE} & \textbf{PerfSAGE} \\
 & \textbf{nn-Meter} & \textbf{(Specialized)} & \textbf{(Generalized)} \\
 \hline
 \emph{CNN-CIFAR10}     & 4.371\% & \textbf{2.314\%} & 2.617\% \\  
 \hline 
 \emph{CNN-ImageNet}    & 3.219\% & \textbf{1.098\%} & 1.263\% \\  
 \hline 
 \emph{CNN-KWS}         & 452.694\% & \textbf{3.722\%} & 5.847\% \\  
 \hline 
 \emph{ViT-CIFAR100}    & --$^\dagger$ & \textbf{4.451\%} & 4.798\% \\  
 \hline 
 \end{tabular}
 }} \quad
 \tiny
 $^\dagger$ Missing nn-Meter's operator support for ViT models prevents characterization.
 \vspace{-3mm}
\end{table}

\label{section:s&g}
Table \ref{table:ete} shows results for \texttt{PerfSAGE} on our \texttt{EdgeDLPerf} dataset. 
The upper section of the table shows the performance of the specialized model for each design space, whereas the lower section shows the performance of the generalized model. 
The generalized model performs slightly worse than the specialized model.
Our model exhibits consistent performance across all available prediction targets on various hardware platforms. Contrary to the findings on nn-Meter~\cite{zhang2021nn}, \texttt{PerfSAGE} prediction performance does not suffer noticeably, even when a specialized hardware accelerator (i.e., NPU or TPU) is used.
The drop in predictive performance for nn-Meter is most likely due to the aggregation function, which assumes that the latency for the entire model is the sum of the latency of each individual kernel. This may be appropriate for hardware such as the CPU, which does not support higher levels of parallelism. This assumption, however, does not hold true when a deep learning hardware accelerator is used for inference, and as a result, their aggregation method results in a reduction in prediction accuracy.

We also applied the nn-Meter methodology to our \texttt{EdgeDLPerf} dataset on a CPU (see Table \ref{table:nnm_comparison}).
nn-Meter is designed primarily for latency prediction; it is unsuitable for other inference performance metrics. 
It also does not fully support 
models from the \emph{ViT-CIFAR100} design space, which require the implementation of a predictor for novel transformer layers.
Some models from \emph{CNN-CIFAR10}, \emph{CNN-ImageNet} and \emph{CNN-KWS} are not suitable for nn-Meter fusion rules. 
Additionally, the prediction error for \emph{CNN-KWS} is far from ideal, due to the requirement for symmetric convolutional kernel shapes.
In contrast, \texttt{PerfSAGE} supports all models from the \texttt{EdgeDLPerf} dataset with lower error rates.  
For a more comprehensive comparison, we adapted the Eagle predictor~\cite{dudziak2020brp} to be able to train it on our dataset.
Although the Eagle predictor is capable of predicting all models in the NASBench201 design space,
it suffers as more model design spaces and hardware platforms are evaluated, as shown in Table \ref{table:ete}.
Eagle shows good performance on NAS-Bench201, partly due to 
prior knowledge of model architecture construction rules. Once that knowledge is hidden from the predictor, as in our benchmark, its accuracy suffers significantly. 
\texttt{PerfSAGE} does not require this information, and nonetheless generalizes well across the diverse landscape. 
Compared to state-of-the-art methods, \texttt{PerfSAGE} provides superior prediction accuracy for more diverse design spaces and prediction targets.

\subsection{Future Work}
The area of learned performance models is a very exciting direction in computer architecture modeling, of which DNN inference performance is merely a narrow aspect.
Learned performance models area a promising direction in the computer architecture modeling field.
We hope that our open source \texttt{EdgeDLPerf} dataset will provide a platform for future improvements on our own \texttt{PerfSAGE} predictor.

More specifically, we envision improvements
in three major areas. 
Firstly, our model still relies on generating a benchmark dataset with a large number of models for training.
Self-supervised training could be applied to our training process to reduce the number of models, which require benchmarking. 
Second, we make predictions only for batch one inference.
In real-world applications, a larger batch size may be used during edge inference.
As a result, we should investigate how to incorporate batch size as additional information for runtime performance prediction.
Third, for use cases such as compiler optimization and Neural Architecture Search, any error in the data learning process will cause a slew of issues in the optimization process.
As a result, it would be useful to associate an uncertainty score with each prediction in order to avoid the pitfalls of erroneous prediction by applying Bayesian Deep Learning methods. 

\section{Conclusion}

\texttt{PerfSAGE} represents an important step toward developing an ideal runtime performance predictor, achieving $<5\%$ MAPE across all prediction targets on three hardware platforms. It  significantly outperforms other state-of-the-art GNN-based methods on \texttt{EdgeDLPerff} dataset. \texttt{PerfSAGE} demonstrates superior generalization ability. Unlike other predictors that do not use a GNN-based model, \texttt{PerfSAGE} can easily adapt to different hardware platforms and prediction targets without the need for manual tuning or implementation changes. It is also highly adaptable to hardware platforms with an optimized kernel-level scheduler. Thus, \texttt{PerfSAGE} overcomes previous works' shortcoming of accuracy loss for prediction of DNN inference performance on deep learning accelerators.

\newpage

\bibliography{References}
\bibliographystyle{mlsys2023}

\appendix

\end{document}